%% file: yolact.tex
\documentclass[10pt,twocolumn,dvipsnames]{article}

\usepackage{template/iccv}
\usepackage{times, epsfig, graphicx, amsmath, amssymb, color}
\usepackage[breaklinks=true,colorlinks,bookmarks=false]{hyperref}

\usepackage[normalem]{ulem}
\usepackage{booktabs, xcolor, subcaption, multirow, tikz}
\usepackage{template/definitions}

\arxivcopy

\def\iccvPaperID{1384}

\begin{document}

\title{\methodname\\Real-time Instance Segmentation}

\author{\vspace{1.5mm} Daniel Bolya \qquad Chong Zhou \qquad Fanyi Xiao \qquad Yong Jae Lee\\
University of California, Davis\\
{\tt\small \{dbolya, cczhou, fyxiao, yongjaelee\}@ucdavis.edu}}

\maketitle
\iccvonly{\thispagestyle{empty}}

\begin{abstract} \vspace*{-0.05in}
    We present a simple, fully-convolutional model for real-time instance segmentation that achieves 29.8 mAP on MS COCO at 33.5 fps evaluated on a single Titan Xp, which is significantly faster than any previous competitive approach. Moreover, we obtain this result after training on \textbf{only one GPU}. We accomplish this by breaking instance segmentation into two parallel subtasks: (1) generating a set of prototype masks and (2) predicting per-instance mask coefficients. Then we produce instance masks by linearly combining the prototypes with the mask coefficients. We find that because this process doesn't depend on repooling, this approach produces very high-quality masks and exhibits temporal stability for free. Furthermore, we analyze the emergent behavior of our prototypes and show they learn to localize instances on their own in a translation variant manner, despite being fully-convolutional. Finally, we also propose Fast NMS, a drop-in 12 ms faster replacement for standard NMS that only has a marginal performance penalty.
\vspace*{-0.05in} \end{abstract}

\vspace*{-0.05in}
\section{Introduction}
	
\emph{``Boxes are stupid anyway though, I'm probably a true believer in masks except I can't get YOLO to learn them.''}
\vspace*{-0.05in}
\begin{flushright}
	    -- Joseph Redmon, YOLOv3~\cite{yolov3}
\end{flushright}

What would it take to create a real-time instance segmentation algorithm? Over the past few years, the vision community has made great strides in instance segmentation, in part by drawing on powerful parallels from the well-established domain of object detection. State-of-the-art approaches to instance segmentation like Mask R-CNN \cite{maskrcnn} and FCIS \cite{fcis} directly build off of advances in object detection like Faster R-CNN \cite{fasterrcnn} and R-FCN \cite{rfcn}. Yet, these methods focus primarily on performance over speed, leaving the scene devoid of instance segmentation parallels to real-time object detectors like SSD \cite{ssd} and YOLO~\cite{yolov2, yolov3}. In this work, our goal is to fill that gap with a fast, one-stage instance segmentation model in the same way that SSD and YOLO fill that gap for object detection.

    \input{figures/speed_performance.tex}

However, instance segmentation is hard---much harder than object detection. One-stage object detectors like SSD and YOLO are able to speed up existing two-stage detectors like Faster R-CNN by simply removing the second stage and making up for the lost performance in other ways. The same approach is not easily extendable, however, to instance segmentation. State-of-the-art two-stage instance segmentation methods depend heavily on \emph{feature localization} to produce masks. That is, these methods ``re-pool'' features in some bounding box region (e.g., via RoI-pool/align), and then feed these now localized features to their mask predictor. This approach is inherently sequential and is therefore difficult to accelerate. One-stage methods that perform these steps in parallel like FCIS do exist, but they require significant amounts of post-processing after localization, and thus are still far from real-time.

To address these issues, we propose \methodname{}\footnote{\textbf{Y}ou \textbf{O}nly \textbf{L}ook \textbf{A}t \textbf{C}oefficien\textbf{T}s}, a real-time instance segmentation framework that forgoes an explicit localization step. Instead, \methodname{} breaks up instance segmentation into two parallel tasks: (1) generating a dictionary of non-local \textit{prototype masks over the entire image}, and (2) predicting a set of \textit{linear combination coefficients per instance}. Then producing a full-image instance segmentation from these two components is simple: for each instance, linearly combine the prototypes using the corresponding predicted coefficients and then crop with a predicted bounding box. We show that by segmenting in this manner, \textit{the network learns how to localize instance masks on its own}, where visually, spatially, and semantically similar instances appear different in the prototypes.

Moreover, since the number of prototype masks is independent of the number of categories (e.g., there can be more categories than prototypes), \methodname{} learns a distributed representation in which each instance is segmented with a combination of prototypes that are shared across categories.  This distributed representation leads to interesting emergent behavior in the prototype space: some prototypes spatially partition the image, some localize instances, some detect instance contours, some encode position-sensitive directional maps (similar to those obtained by hard-coding a position-sensitive module in FCIS~\cite{fcis}), and most do a combination of these tasks (see Figure~\ref{fig:behavior}).

This approach also has several practical advantages. First and foremost, it's fast: because of its parallel structure and extremely lightweight assembly process, \methodname{} adds only a marginal amount of computational overhead to a one-stage backbone detector, making it easy to reach 30 fps even when using ResNet-101 \cite{resnet}; in fact, \emph{the entire mask branch takes only $\sim$5 ms to evaluate}. Second, masks are high-quality: since the masks use the full extent of the image space without any loss of quality from repooling, our masks for large objects are significantly higher quality than those of other methods (see Figure~\ref{fig:mask_quality}). Finally, it's general: the idea of generating prototypes and mask coefficients could be added to almost any modern object detector.


Our main contribution is the first real-time ($>30$ fps) instance segmentation algorithm with competitive results on the challenging MS COCO dataset~\cite{coco} (see Figure~\ref{fig:speed_performance}). In addition, we analyze the emergent behavior of \methodname{}'s prototypes and provide experiments to study the speed vs.~performance trade-offs obtained with different backbone architectures, numbers of prototypes, and image resolutions. We also provide a novel Fast NMS approach that is 12ms faster than traditional NMS with a negligible performance penalty. The code for YOLACT is available at \url{https://github.com/dbolya/yolact}.

\section{Related Work}
\vspace*{-0.1in}

\paragraph{Instance Segmentation}
Given its importance, a lot of research effort has been made to push instance segmentation \emph{accuracy}. Mask-RCNN~\cite{maskrcnn} is a representative two-stage instance segmentation approach that first generates candidate region-of-interests (ROIs) and then classifies and segments those ROIs in the second stage. Follow-up works try to improve its accuracy by e.g., enriching the FPN features~\cite{liu-panet2018} or addressing the incompatibility between a mask's confidence score and its localization accuracy~\cite{huang-msrcnn2018}. These two-stage methods require re-pooling features for each ROI and processing them with subsequent computations, which make them unable to obtain real-time speeds (30 fps) even when decreasing image size (see Table \ref{tab:accelerated_baselines}).

One-stage instance segmentation methods generate position sensitive maps that are assembled into final masks with position-sensitive pooling~\cite{dai-eccv2016,fcis} or combine semantic segmentation logits and direction prediction logits~\cite{chen-masklab2018}. Though conceptually faster than two-stage methods, they still require repooling or other non-trivial computations (e.g.,~mask voting). This severely limits their speed, placing them far from real-time. In contrast, our assembly step is much more lightweight (only a linear combination) and can be implemented as one GPU-accelerated matrix-matrix multiplication, making our approach very fast.

Finally, some methods first perform semantic segmentation followed by boundary detection~\cite{kirillov-cvpr2017}, pixel clustering~\cite{bai-cvpr2017,liang-pami2018}, or learn an embedding to form instance masks~\cite{newell-nips2017,harley-iccv2017,de-arxiv2017,fathi-arxiv2017}. Again, these methods have multiple stages and/or involve expensive clustering procedures, which limits their viability for real-time applications.

    \input{figures/concept.tex}

\paragraph{Real-time Instance Segmentation}
While real-time object detection~\cite{ssd,yolov1,yolov2,yolov3}, and semantic segmentation~\cite{segnet,treml2016sq,enet,blitznet,zhao2018icnet} methods exist, few works have focused on real-time instance segmentation.  Straight to Shapes~\cite{straight2shapes} and Box2Pix~\cite{box2pix} can perform instance segmentation in real-time (30 fps on Pascal SBD 2012~\cite{pascalvoc,sbd} for Straight to Shapes, and 10.9 fps on Cityscapes~\cite{cityscape} and 35 fps on KITTI~\cite{KITTI} for Box2Pix), but their accuracies are far from that of modern baselines. In fact, Mask R-CNN~\cite{maskrcnn} remains one of the fastest instance segmentation methods on semantically challenging datasets like COCO~\cite{coco} (13.5 fps on $550^2$ px images; see Table~\ref{tab:accelerated_baselines}).


\paragraph{Prototypes}
Learning prototypes (aka vocabulary or codebook) has been extensively explored in computer vision. Classical representations include textons~\cite{textons} and visual words~\cite{sivic-iccv2003}, with advances made via sparsity and locality priors~\cite{yang-tip2010,wang-cvpr2010,zhang-iccv2013}. Others have designed prototypes for object detection~\cite{agarwal-eccv2002,yu-bmvc2007,ren-cvpr2013}. Though related, these works use prototypes to represent features, whereas we use them to assemble masks for instance segmentation. Moreover, we learn prototypes that are specific to each image, rather than global prototypes shared across the entire dataset.

\section{\methodname}

Our goal is to add a mask branch to an existing one-stage object detection model in the same vein as Mask R-CNN \cite{maskrcnn} does to Faster R-CNN \cite{fasterrcnn}, but without an explicit feature localization step (e.g., feature repooling). To do this, we break up the complex task of instance segmentation into two simpler, parallel tasks that can be assembled to form the final masks. The first branch uses an FCN \cite{fcn} to produce a set of image-sized ``prototype masks'' that do not depend on any one instance. The second adds an extra head to the object detection branch to predict a vector of ``mask coefficients'' for each anchor that encode an instance's representation in the prototype space. Finally, for each instance that survives NMS, we construct a mask for that instance by linearly combining the work of these two branches.

\paragraph{Rationale}
We perform instance segmentation in this way primarily because masks are spatially coherent; i.e., pixels close to each other are likely to be part of the same instance. While a convolutional (\textit{conv}) layer naturally takes advantage of this coherence, a fully-connected (\textit{fc}) layer does not. That poses a problem, since one-stage object detectors produce class and box coefficients for each anchor as an output of an \textit{fc} layer.\footnotemark ~Two stage approaches like Mask R-CNN get around this problem by using a localization step (e.g., RoI-Align), which preserves the spatial coherence of the features while also allowing the mask to be a \textit{conv} layer output. However, doing so requires a significant portion of the model to wait for a first-stage RPN to propose localization candidates, inducing a significant speed penalty.

\footnotetext{To show that this is an issue, we develop an ``\textit{fc}-mask'' model that produces masks for each anchor as the reshaped output of an \textit{fc} layer. As our experiments in Table \ref{tab:accelerated_baselines} show, simply adding masks to a one-stage model as \textit{fc} outputs only obtains 20.7 mAP and is thus very much insufficient.}

Thus, we break the problem into two parallel parts, making use of \textit{fc} layers, which are good at producing semantic vectors, and \textit{conv} layers, which are good at producing spatially coherent masks, to produce the ``mask coefficients'' and ``prototype masks'', respectively. Then, because prototypes and mask coefficients can be computed independently, the computational overhead over that of the backbone detector comes mostly from the assembly step, which can be implemented as a single matrix multiplication.  In this way, we can maintain spatial coherence in the feature space while still being one-stage and \emph{fast}.

\subsection{Prototype Generation}
\vspace*{-0.01in}
The prototype generation branch (protonet) predicts a set of $k$ prototype masks for the entire image. We implement protonet as an FCN whose last layer has $k$ channels (one for each prototype) and attach it to a backbone feature layer (see~Figure~\ref{fig:protonet} for an illustration). While this formulation is similar to standard semantic segmentation, it differs in that we exhibit no explicit loss on the prototypes. Instead, all supervision for these prototypes comes from the final mask loss after assembly.

We note two important design choices: taking protonet from deeper backbone features produces more robust masks, and higher resolution prototypes result in both higher quality masks and better performance on smaller objects. Thus, we use FPN \cite{fpn} because its largest feature layers ($P_3$ in our case; see Figure~\ref{fig:concept}) are the deepest. Then, we upsample it to one fourth the dimensions of the input image to increase performance on small objects.

    \input{figures/protonet}
    
Finally, we find it important for the protonet's output to be unbounded, as this allows the network to produce large, overpowering activations for prototypes it is very confident about (e.g., obvious background). Thus, we have the option of following protonet with either a {\tt ReLU} or no nonlinearity. We choose {\tt ReLU} for more interpretable prototypes.

\subsection{Mask Coefficients}
Typical anchor-based object detectors have two branches in their prediction heads: one branch to predict $c$ class confidences, and the other to predict $4$ bounding box regressors. For mask coefficient prediction, we simply add a third branch in parallel that predicts $k$ mask coefficients, one corresponding to each prototype. Thus, instead of producing $4 + c$ coefficients per anchor, we produce $4 + c + k$.

Then for nonlinearity, we find it important to be able to subtract out prototypes from the final mask. Thus, we apply {\tt tanh} to the $k$ mask coefficients, which produces more stable outputs over no nonlinearity. The relevance of this design choice is apparent in Figure \ref{fig:concept}, as neither mask would be constructable without allowing for subtraction.

\subsection{Mask Assembly}
\vspace{-0.01in}
To produce instance masks, we combine the work of the prototype branch and mask coefficient branch, using a linear combination of the former with the latter as coefficients. We then follow this by a sigmoid nonlinearity to produce the final masks. These operations can be implemented efficiently using a single matrix multiplication and sigmoid:
    \begin{align} \vspace{-0.3in}
        M = \sigma{(P C^{T})} 
    \vspace{-0.2in} \end{align}
where $P$ is an $h\times w \times k$ matrix of prototype masks and $C$ is a $n \times k$ matrix of mask coefficients for $n$ instances surviving NMS and score thresholding. Other, more complicated combination steps are possible; however, we keep it simple (and fast) with a basic linear combination.

    \input{figures/pred_head}

\paragraph{Losses} We use three losses to train our model: classification loss $L_{cls}$, box regression loss $L_{box}$ and mask loss $L_{mask}$ with the weights 1, 1.5, and 6.125 respectively. Both $L_{cls}$ and $L_{box}$ are defined in the same way as in~\cite{ssd}. Then to compute mask loss, we simply take the pixel-wise binary cross entropy between assembled masks $M$ and the ground truth masks $M_{gt}$: $L_{mask} = \text{BCE}(M, M_{gt})$.

\paragraph{Cropping Masks}
\label{sec:crop_mask}
We crop the final masks with the predicted bounding box during evaluation.  During training, we instead crop with the ground truth bounding box, and divide $L_{mask}$ by the ground truth bounding box area to preserve small objects in the prototypes.

\subsection{Emergent Behavior}
Our approach might seem surprising, as the general consensus around instance segmentation is that because FCNs are translation invariant, the task needs translation variance added back in \cite{fcis}. Thus methods like FCIS \cite{fcis} and Mask R-CNN \cite{maskrcnn} try to explicitly add translation variance, whether it be by directional maps and position-sensitive repooling, or by putting the mask branch in the second stage so it does not have to deal with localizing instances. In our method, the only translation variance we add is to crop the final mask with the predicted bounding box. However, we find that our method also works without cropping for medium and large objects, so this is not a result of cropping. Instead, \methodname{} \textit{learns how to localize instances on its own} via different activations in its prototypes.

To see how this is possible, first note that the prototype activations for the solid red image (image a) in Figure \ref{fig:behavior} are actually not possible in an FCN without padding. Because a convolution outputs to a single pixel, if its input everywhere in the image is the same, the result everywhere in the \textit{conv} output will be the same. On the other hand, the consistent rim of padding in modern FCNs like ResNet gives the network the ability to tell how far away from the image's edge a pixel is. Conceptually, one way it could accomplish this is to have multiple layers in sequence spread the padded 0's out from the edge toward the center (e.g., with a kernel like $[1, 0]$). This means ResNet, for instance, \textit{is inherently translation variant}, and our method makes heavy use of that property (images b and c exhibit clear translation variance).

We observe many prototypes to activate on certain ``partitions'' of the image. That is, they only activate on objects on one side of an implicitly learned boundary. In Figure \ref{fig:behavior}, prototypes 1-3 are such examples. By combining these partition maps, the network can distinguish between different (even overlapping) instances of the same semantic class; e.g., in image d, the green umbrella can be separated from the red one by subtracting prototype 3 from prototype 2.

    \input{figures/behavior.tex}

Furthermore, being learned objects, prototypes are compressible. That is, if protonet combines the functionality of multiple prototypes into one, the mask coefficient branch can learn which situations call for which functionality. For instance, in Figure \ref{fig:behavior}, prototype 2 is a partitioning prototype but also fires most strongly on instances in the bottom-left corner. Prototype 3 is similar but for instances on the right. This explains why in practice, the model does not degrade in performance even with as low as $k=32$ prototypes (see Table~\ref{tab:num_proto}).
On the other hand, increasing $k$ is ineffective most likely because predicting coefficients is difficult. If the network makes a large error in even one coefficient, due to the nature of linear combinations, the produced mask can vanish or include leakage from other objects. Thus, the network has to play a balancing act to produce the right coefficients, and adding more prototypes makes this harder. In fact, we find that for higher values of $k$, the network simply adds redundant prototypes with small edge-level variations that slightly increase $AP_{95}$, but not much else.

\section{Backbone Detector}

For our backbone detector we prioritize speed as well as feature richness, since predicting these prototypes and coefficients is a difficult task that requires good features to do well. Thus, the design of our backbone detector closely follows RetinaNet \cite{retinanet} with an emphasis on speed.

\paragraph{\methodname{} Detector}
We use ResNet-101 \cite{resnet} with FPN \cite{fpn} as our default feature backbone and a base image size of $550 \times 550$. We do not preserve aspect ratio in order to get consistent evaluation times per image. Like RetinaNet, we modify FPN by not producing $P_2$ and producing $P_6$ and $P_7$ as successive $3 \times 3$ stride 2 \textit{conv} layers starting from $P_5$ (not $C_5$) and place 3 anchors with aspect ratios $[1, 1/2, 2]$ on each. The anchors of $P_3$ have areas of $24$ pixels squared, and every subsequent layer has double the scale of the previous (resulting in the scales $[24, 48, 96, 192, 384]$). For the prediction head attached to each $P_i$, we have one $3 \times 3$ \textit{conv} shared by all three branches, and then each branch gets its own $3 \times 3$ \text{conv} in parallel. Compared to RetinaNet, our prediction head design (see Figure \ref{fig:pred_head}) is more lightweight and much faster. We apply smooth-$L_1$ loss to train box regressors and encode box regression coordinates in the same way as SSD \cite{ssd}. To train class prediction, we use softmax cross entropy with $c$ positive labels and 1 background label, selecting training examples using OHEM \cite{ohem} with a 3:1 neg:pos ratio. Thus, unlike RetinaNet we do not use focal loss, which we found not to be viable in our situation.

With these design choices, we find that this backbone performs better and faster than SSD \cite{ssd} modified to use ResNet-101 \cite{resnet}, with the same image size.

\section{Other Improvements}
We also discuss other improvements that either increase speed with little effect on performance or increase performance with no speed penalty.

    \input{figures/qualitative.tex}

\paragraph{Fast NMS}
After producing bounding box regression coefficients and class confidences for each anchor, like most object detectors we perform NMS to suppress duplicate detections. In many previous works \cite{yolov2, yolov3, ssd, fasterrcnn, maskrcnn, retinanet}, NMS is performed sequentially. That is, for each of the $c$ classes in the dataset, sort the detected boxes descending by confidence, and then for each detection remove all those with lower confidence than it that have an IoU overlap greater than some threshold. While this sequential approach is fast enough at speeds of around 5 fps, it becomes a large barrier for obtaining 30 fps (for instance, a 10 ms improvement at 5 fps results in a 0.26 fps boost, while a 10 ms improvement at 30 fps results in a 12.9 fps boost).

To fix the sequential nature of traditional NMS, we introduce Fast NMS, a version of NMS where every instance can be decided to be kept or discarded in parallel. To do this, we simply allow already-removed detections to suppress other detections, which is not possible in traditional NMS. This relaxation allows us to implement Fast NMS entirely in standard GPU-accelerated matrix operations.

To perform Fast NMS, we first compute a $c \times n\times n$ pairwise IoU matrix $X$ for the top $n$ detections sorted descending by score for each of $c$ classes. Batched sorting on the GPU is readily available and computing IoU can be easily vectorized. Then, we remove detections if there are any higher-scoring detections with a corresponding IoU greater than some threshold $t$. We efficiently implement this by first setting the lower triangle and diagonal of $X$ to 0: $X_{kij} = 0, ~\forall k, j, i \geq j,$
which can be performed in one batched {\tt triu} call, and then taking the column-wise max:
    \begin{align} \label{eq:max}
        K_{kj} = \max_i(X_{kij}) \qquad \forall k, j
    \end{align} 
to compute a matrix $K$ of maximum IoU values for each detection. Finally, thresholding this matrix with $t$ ($K < t$) will indicate which detections to keep for each class.

Because of the relaxation, Fast NMS has the effect of removing slightly too many boxes. However, the performance hit caused by this is negligible compared to the stark increase in speed (see Table \ref{tab:nms}). In our code base, Fast NMS is 11.8 ms faster than a Cython implementation of traditional NMS while only reducing performance by 0.1 mAP. In the Mask R-CNN benchmark suite \cite{maskrcnn}, Fast NMS is 15.0 ms faster than their CUDA implementation of traditional NMS with a performance loss of only 0.3 mAP.

    \input{figures/mask_quality.tex}

\vspace{-0.01in}
\paragraph{Semantic Segmentation Loss}
While Fast NMS trades a small amount of performance for speed, there are ways to increase performance with no speed penalty. One of those ways is to apply extra losses to the model during training using modules not executed at test time. This effectively increases feature richness while at no speed penalty.

Thus, we apply a semantic segmentation loss on our feature space using layers that are only evaluated during training. Note that because we construct the ground truth for this loss from instance annotations, this does not strictly capture semantic segmentation (i.e., we do not enforce the standard one class per pixel).
To create predictions during training, we simply attach a 1x1 $\textit{conv}$ layer with $c$ output channels directly to the largest feature map ($P_3$) in our backbone. Since each pixel can be assigned to more than one class, we use sigmoid and $c$ channels instead of softmax and $c+1$. This loss is given a weight of 1 and results in a $+0.4$ mAP boost.

\vspace{-0.05in}
\section{Results} \label{sec:results}
\vspace{-0.05in}

We report instance segmentation results on MS COCO \cite{coco} and Pascal 2012 SBD~\cite{sbd} using the standard metrics. For MS COCO, we train on {\tt train2017} and evaluate on {\tt val2017} and {\tt test-dev}.

\vspace{-0.01in}
\paragraph{Implementation Details}
We train all models with batch size 8 \textit{on one GPU} using ImageNet \cite{imagenet} pretrained weights. We find that this is a sufficient batch size to use batch norm, so we leave the pretrained batch norm unfrozen but do not add any extra \textit{bn} layers. We train with SGD for 800k iterations starting at an initial learning rate of $10^{-3}$ and divide by 10 at iterations 280k, 600k, 700k, and 750k, using a weight decay of $5{\times}10^{-4}$, a momentum of 0.9, and all data augmentations used in SSD \cite{ssd}. For Pascal, we train for 120k iterations and divide the learning rate at 60k and 100k. We also multiply the anchor scales by $4/3$, as objects tend to be larger. Training takes 4-6 days (depending on config) on one Titan Xp for COCO and less than 1 day on Pascal.

\paragraph{Mask Results}
We first compare \methodname{} to state-of-the art methods on MS COCO's {\tt test-dev} set in Table \ref{tab:performance}. Because our main goal is speed, we compare against other single model results with no test-time augmentations. We report all speeds computed on a single Titan Xp, so some listed speeds may be faster than in the original paper.

    \input{figures/performance.tex}
    \input{figures/ablations.tex}
    \input{figures/pascal.tex}

\methodname{}-550 offers competitive instance segmentation performance while at 3.8x the speed of the previous fastest instance segmentation method on COCO. We also note an interesting difference in where the performance of our method lies compared to others. Supporting our qualitative findings in Figure~\ref{fig:mask_quality}, the gap between \methodname{}-550 and Mask R-CNN at the 50\% overlap threshold is 9.5 AP, while it's 6.6 at the 75\% IoU threshold. This is different from the performance of FCIS, for instance, compared to Mask R-CNN where the gap is consistent (AP values of 7.5 and 7.6 respectively). Furthermore, at the highest (95\%) IoU threshold, we outperform Mask R-CNN with 1.6 vs.~1.3 AP.

We also report numbers for alternate model configurations in Table~\ref{tab:performance}. In addition to our base $550\times 550$ image size model, we train $400 \times 400$ (\methodname{}-400) and $700 \times 700$ (\methodname{}-700) models, adjusting the anchor scales accordingly ($s_x = s_{550} / 550 * x$). Lowering the image size results in a large decrease in performance, demonstrating that instance segmentation naturally demands larger images. Then, raising the image size decreases speed significantly but also increases performance, as expected.

In addition to our base backbone of ResNet-101 \cite{resnet}, we also test ResNet-50 and DarkNet-53 \cite{yolov3} to obtain even faster results. If higher speeds are preferable we suggest using ResNet-50 or DarkNet-53 instead of lowering the image size, as these configurations perform much better than \methodname{}-400, while only being slightly slower.

Finally, we also train and evaluate our ResNet-50 model on Pascal 2012 SBD in Table~\ref{tab:pascal}. YOLACT clearly outperforms popular approaches that report SBD performance, while also being significantly faster.

\paragraph{Mask Quality} 
Because we produce a final mask of size $138 \times 138$, and because we create masks directly from the original features (with no repooling to transform and potentially misalign the features), our masks for large objects are noticeably higher quality than those of Mask R-CNN \cite{maskrcnn} and FCIS \cite{fcis}. For instance, in Figure~\ref{fig:mask_quality}, \methodname{} produces a mask that cleanly follows the boundary of the arm, whereas both FCIS and Mask R-CNN have more noise. Moreover, despite being 5.9 mAP worse overall, at the 95\% IoU threshold, our base model achieves 1.6 AP while Mask R-CNN obtains 1.3. This indicates that repooling does result in a quantifiable decrease in mask quality.

\paragraph{Temporal Stability}
Although we only train using static images and do not apply any temporal smoothing, we find that our model produces more temporally stable masks on videos than Mask R-CNN, whose masks jitter across frames even when objects are stationary. We believe our masks are more stable in part because they are higher quality (thus there is less room for error between frames), but mostly because our model is one-stage. Masks produced in two-stage methods are highly dependent on their region proposals in the first stage. In contrast for our method, even if the model predicts different boxes across frames, the prototypes are not affected, yielding much more temporally stable masks. \reviewonly{Please see the supplementary details for real-time video results.}


    \arxivonly{\input{figures/more_qualitative.tex}}

\vspace{-0.05in}
\section{Discussion}
\vspace{-0.05in}

Despite our masks being higher quality and having nice properties like temporal stability, we fall behind state-of-the-art instance segmentation methods in overall performance, albeit while being much faster. Most errors are caused by mistakes in the detector: misclassification, box misalignment, etc. However, we have identified two typical errors caused by \methodname{}'s mask generation algorithm.

\vspace{-0.05in}
\paragraph{Localization Failure}
If there are too many objects in one spot in a scene, the network can fail to localize each object in its own prototype. In these cases, the network will output something closer to a foreground mask than an instance segmentation for some objects in the group; e.g., in the first image in Figure~\ref{fig:qualitative} (row 1 column 1), the blue truck under the red airplane is not properly localized.

\vspace{-0.05in}
\paragraph{Leakage}
Our network leverages the fact that masks are cropped after assembly, and makes no attempt to suppress noise outside of the cropped region. This works fine when the bounding box is accurate, but when it is not, that noise can creep into the instance mask, creating some ``leakage'' from outside the cropped region. This can also happen when two instances are far away from each other, because the network has learned that it doesn't need to localize far away instances---the cropping will take care of it. However, if the predicted bounding box is too big, the mask will include some of the far away instance's mask as well. For instance, Figure~\ref{fig:qualitative} (row 2 column 4) exhibits this leakage because the mask branch deems the three skiers to be far enough away to not have to separate them.
 

\vspace{-0.05in}
\paragraph{Understanding the AP Gap}
However, localization failure and leakage alone are not enough to explain the almost 6 mAP gap between YOLACT's base model and, say, Mask R-CNN. Indeed, our base model on COCO has just a 2.5 mAP difference between its test-dev mask and box mAP (29.8 mask, 32.3 box), meaning our base model would only gain a few points of mAP even with perfect masks. Moreover, Mask R-CNN has this same mAP difference (35.7 mask, 38.2 box), which suggests that the gap between the two methods lies in the relatively poor performance of our detector and not in our approach to generating masks.


\vspace{-0.05in}
\paragraph{Acknowledgements}
This work was supported in part by ARO YIP W911NF17-1-0410, NSF CAREER IIS-1751206, AWS ML Research Award, Google Cloud Platform research credits, and XSEDE IRI180001. 

\appendix
\section{Appendix}
\subsection{Box Results}
    Since \methodname{} produces boxes in addition to masks, we can also compare its object detection performance to other real-time object detection methods. Moreover, while our \textit{mask performance} is real-time, we don't need to produce masks to run \methodname{} as an object detector. Thus, \methodname{} is faster when run to produce boxes than when run to produce instance segmentations. In Table~\ref{tab:detection}, we compare our performance and speed to various skews of YOLOv3 \cite{yolov3}.

    \arxivonly{\input{figures/detection.tex}}
    
    Thus, we are able to achieve similar detection results to YOLOv3 at similar speeds, while not employing any of the additional improvements in YOLOv2 and YOLOv3 like multi-scale training, optimized anchor boxes, cell-based regression encoding, and objectness score. Because the improvements to our detection performance in our observation come mostly from using FPN and training with masks (both of which are orthogonal to the improvements that YOLO makes), it is likely that we can combine YOLO and \methodname{} to create an even better detector.

    Moreover, these detection results show that our mask branch takes \emph{only 6 ms} in total to evaluate, which demonstrates how minimal our mask computation is.

\subsection{More Qualitative Results}
    Figure~\ref{fig:qualitative} shows many examples of adjacent people and vehicles, but not many for other classes. To further support that \methodname{} is not just doing semantic segmentation, we include many more qualitative results for images with adjacent instances of the same class in Figure~\ref{fig:more_qualitative}.
    
    For instance, in an image with two elephants (Figure~\ref{fig:more_qualitative} row 2, col 2), despite the fact that two instance boxes are overlapping with each other, their masks are clearly separating the instances. This is also clearly manifested in the examples of zebras (row 4, col 2) and birds (row 5, col 1). 
    
    Note that for some of these images, the box doesn't exactly crop off the mask. This is because for speed reasons (and because the model was trained in this way), we crop the mask at the prototype resolution (so one fourth the image resolution) with 1px of padding in each direction. On the other hand, the corresponding box is displayed at the original image resolution with no padding.

{\small
    \bibliographystyle{template/ieee_fullname}
    \bibliography{yolact}
}

\end{document}

%% file: figures/speed_performance.tex
\begin{figure}[t!]
    \centering
    \hspace*{-0.5cm} \includegraphics[trim=5 0 0 0, clip, width=0.43\textwidth]{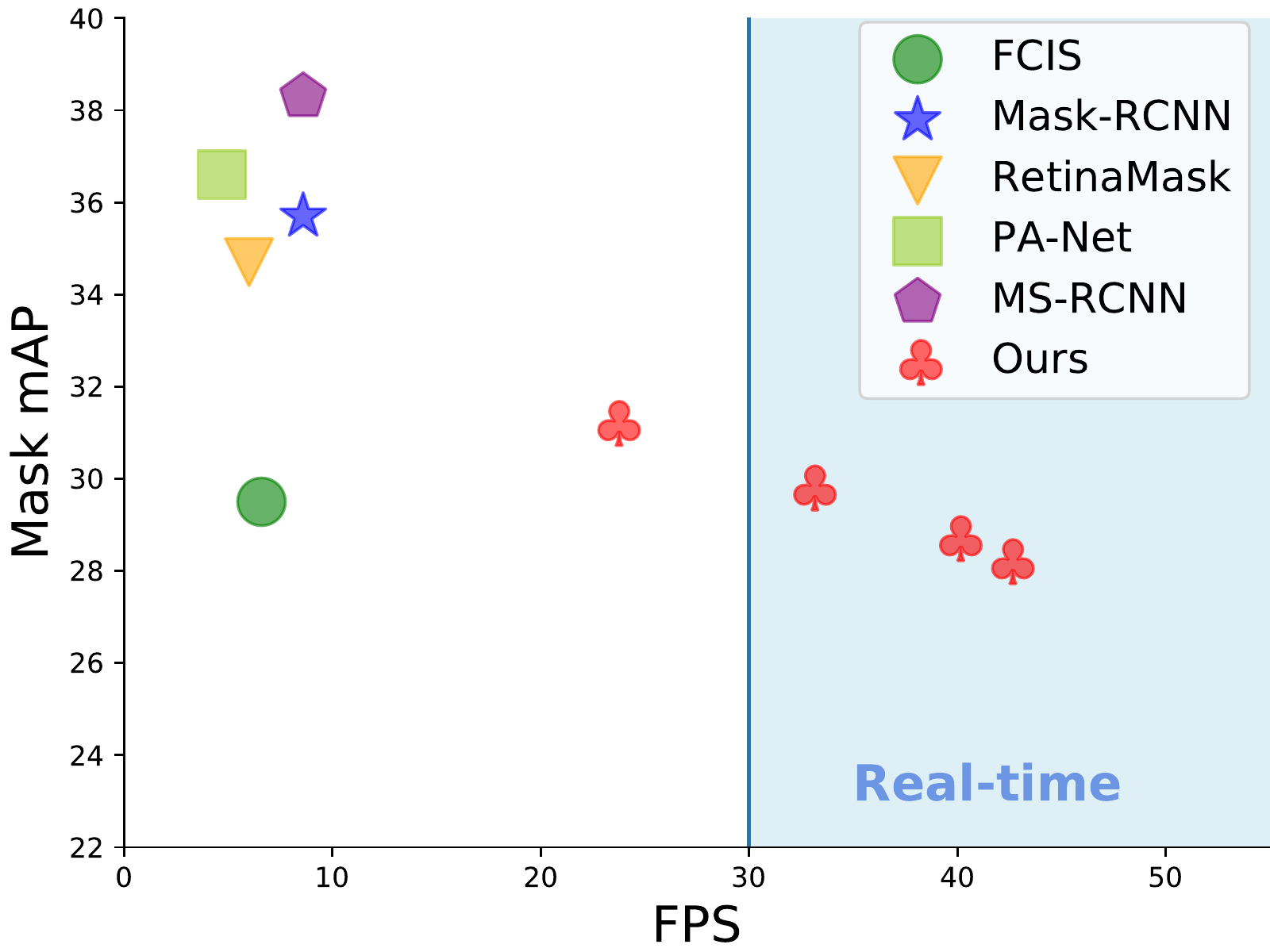}
    \vspace{-0.12in}
    \caption{Speed-performance trade-off for various instance segmentation methods on COCO. To our knowledge, ours is the first \emph{real-time} (above 30 FPS) approach with around 30 mask mAP on COCO {\tt test-dev}. }
    \vspace{-0.13in}
    \label{fig:speed_performance}
\end{figure}

%% file: figures/concept.tex
\begin{figure*}
    \centering
    \includegraphics[trim=10 0 5 180, clip, width=0.98\textwidth]{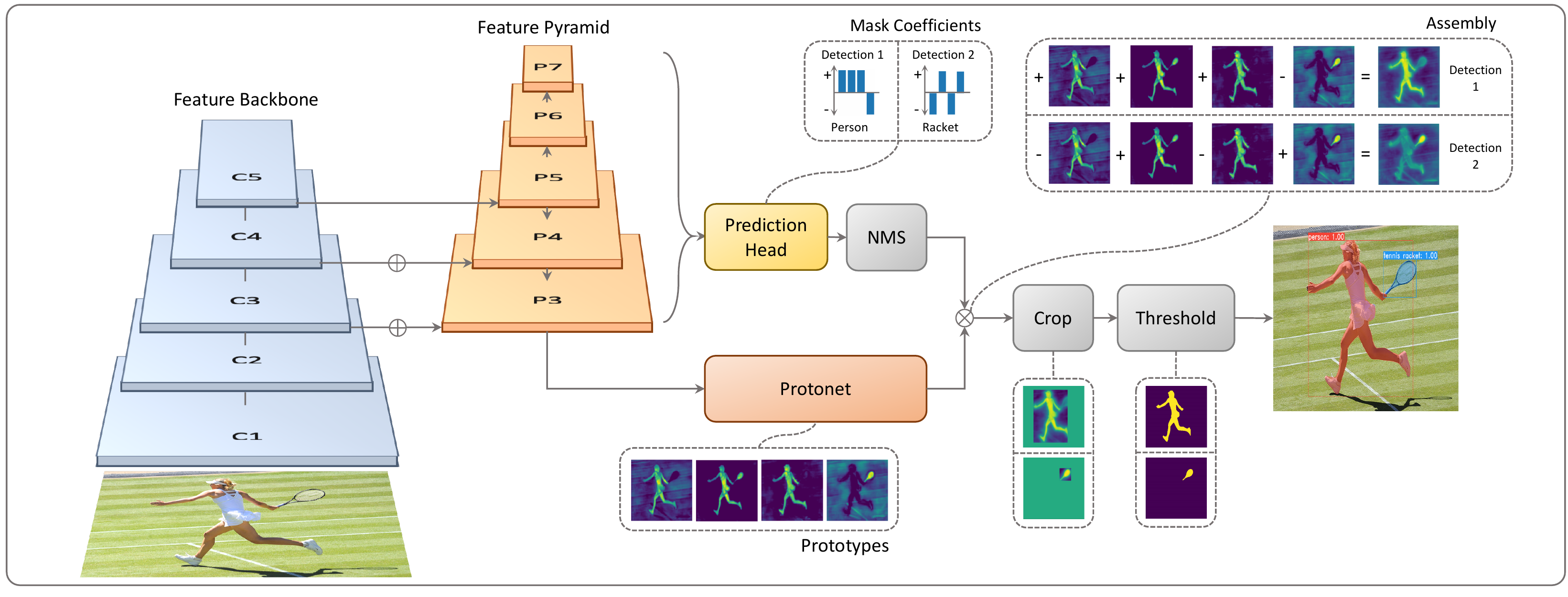}
    \vspace{-0.14in}
    \caption{\inlinesection{\methodname{} Architecture} Blue/yellow indicates low/high values in the prototypes, gray nodes indicate functions that are not trained, and $k=4$ in this example. We base this architecture off of RetinaNet~\cite{retinanet} using ResNet-101 + FPN.}
    \vspace{-0.11in}
    \label{fig:concept}
\end{figure*}

%% file: figures/protonet.tex
\begin{figure}
    \centering
    
    \includegraphics[trim=135 230 230 190, clip, scale=0.35]{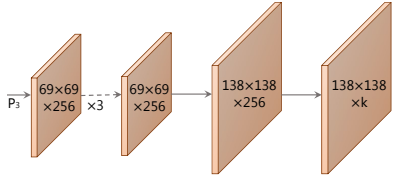}
    \vspace{-0.06in}
    
    \caption{\inlinesection{Protonet Architecture} The labels denote feature size and channels for an image size of $550 \times 550$. Arrows indicate $3\times 3$ \textit{conv} layers, except for the final \textit{conv} which is $1 \times 1$. The increase in size is an upsample followed by a \textit{conv}. Inspired by the mask branch in \cite{maskrcnn}. }
    \vspace{-0.1in}
    \label{fig:protonet}
\end{figure}

%% file: figures/pred_head.tex
\begin{figure}
    \centering
    
    \begin{tikzpicture}
        \draw (0, 0) node [inner sep=0] {\includegraphics[trim=20 85 40 75, clip, scale=0.32]{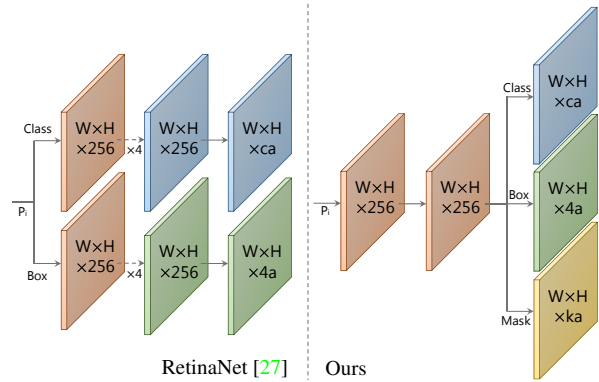}};
        \draw (-0.3, -2.4) node [anchor=east, font=\footnotesize] {RetinaNet \cite{retinanet}};
        \draw (-0.05, -2.4) node [anchor=west, font=\footnotesize] {Ours};
    \end{tikzpicture}
    \vspace{-0.26in}
    \caption{\inlinesection{Head Architecture} We use a shallower prediction head than RetinaNet~\cite{retinanet} and add a mask coefficient branch. This is for $c$ classes, $a$ anchors for feature layer $P_i$, and $k$ prototypes. See Figure~\ref{fig:protonet} for a key.}
    \vspace{-0.11in}
    \label{fig:pred_head}
\end{figure}

%% file: figures/behavior.tex
\begin{figure}[t!]
    \centering
    \includegraphics[width=0.46\textwidth]{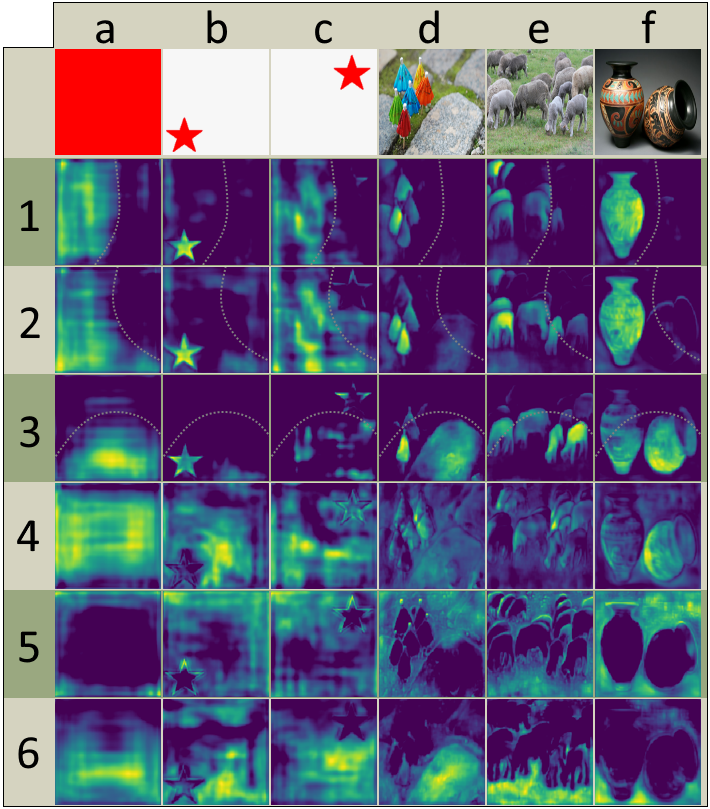}\vspace{-0.1in}
    \caption{\inlinesection{Prototype Behavior} The activations of the same six prototypes (y axis) across different images (x axis). Prototypes 1-3 respond to objects to one side of a soft, implicit boundary (marked with a dotted line). Prototype 4 activates on the bottom-left of objects (for instance, the bottom left of the umbrellas in image d); prototype 5 activates on the background and on the edges between objects; and prototype 6 segments what the network perceives to be the ground in the image. These last 3 patterns are most clear in images d-f.}
    \vspace{-0.11in}
    \label{fig:behavior}
    
\end{figure}

%% file: figures/qualitative.tex
\begin{figure*}
    \centering
    \includegraphics[width=0.9\textwidth]{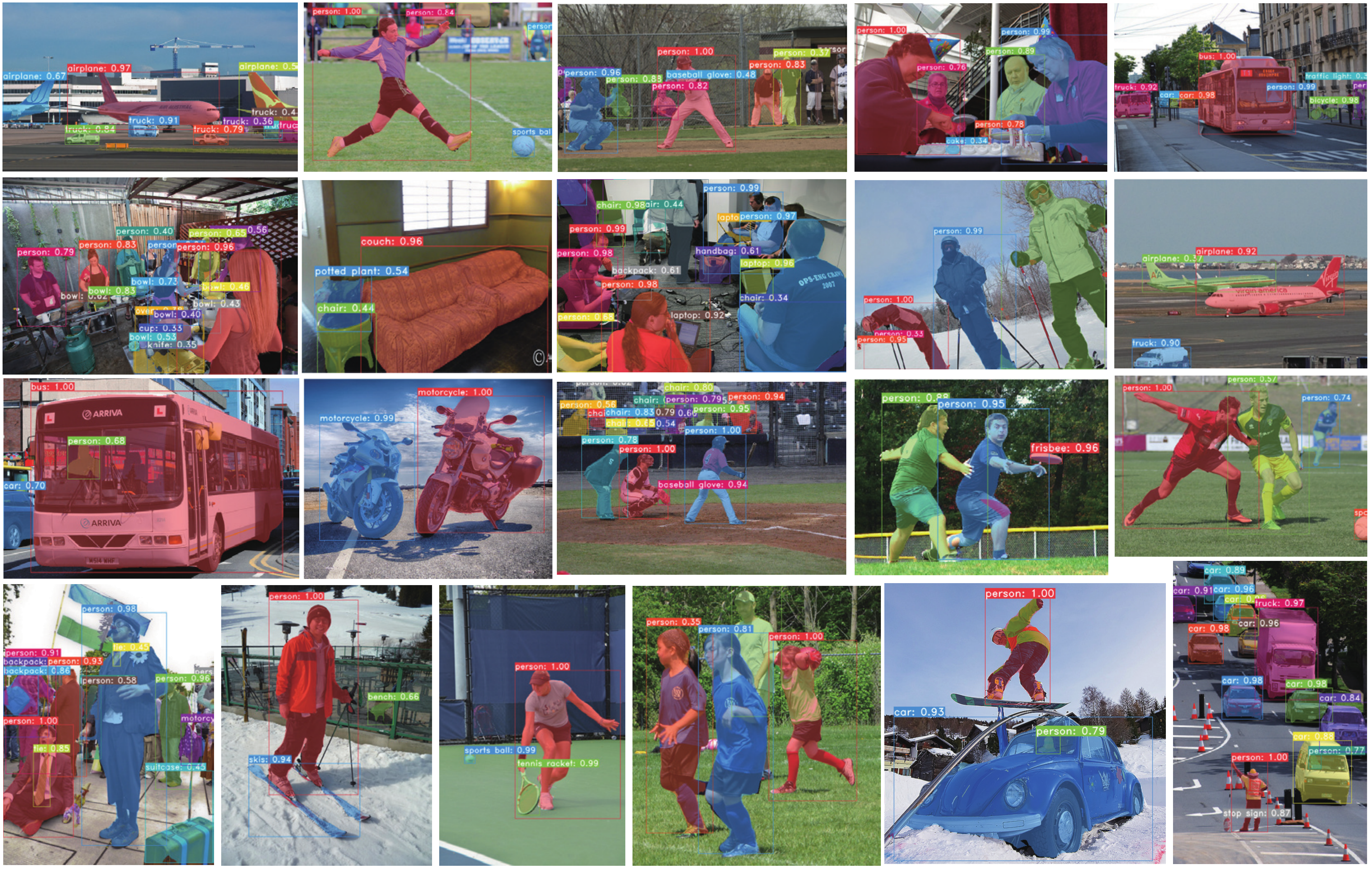}
    \vspace{-0.12in}
    \caption{\textbf{\methodname} evaluation results on COCO's \texttt{test-dev} set. This base model achieves 29.8 mAP at 33.0 fps. All images have the confidence threshold set to 0.3. \reviewonly{Please see the supplementary details for real-time video results.}}
    \vspace{-0.15in}
    \label{fig:qualitative}
\end{figure*}

%% file: figures/mask_quality.tex
\begin{figure*}[t!]
    \centering
    \includegraphics[width=0.97\textwidth]{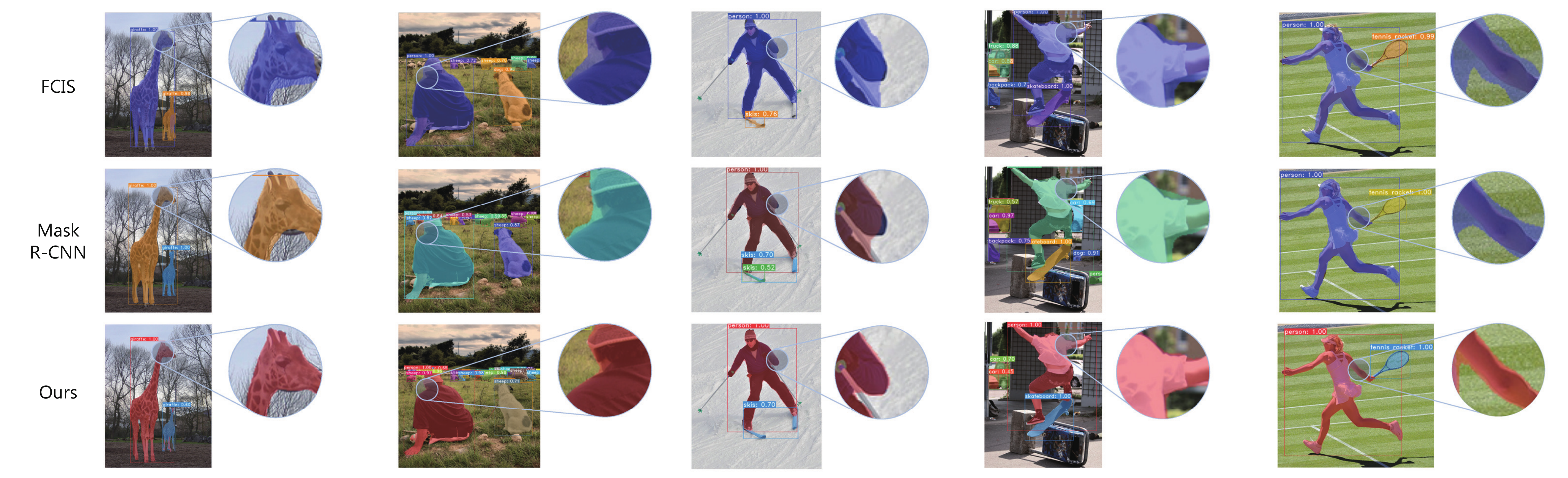}\vspace{-0.13in}
    \caption{\inlinesection{Mask Quality} Our masks are typically higher quality than those of Mask R-CNN \cite{maskrcnn} and FCIS \cite{fcis} because of the larger mask size and lack of feature repooling.}
    \vspace{-0.13in}
    \label{fig:mask_quality}
\end{figure*}

%% file: figures/performance.tex
\begin{table*}
    \centering
    
    \def\mrcnn{Mask R-CNN \cite{maskrcnn}}
    \def\fcis{FCIS \cite{fcis}}
    \def\panet{PA-Net~\cite{liu-panet2018}}
    \def\msrcnn{MS R-CNN~\cite{huang-msrcnn2018}}
    \def\retinamask{RetinaMask~\cite{fu-retinamask2019}}
    \def\masklab{MaskLab~\cite{chen-masklab2018}}

    \newcommand{\modelname}[1]{\methodname{}-#1}
    
    \begin{smalltable}{l c l cc ccc c ccc cc} \toprule
        Method          && Backbone  &&    FPS    &    Time   &&    AP     & AP$_{50}$ & AP$_{75}$ &&  AP$_{S}$ &  AP$_{M}$ &  AP$_{L}$ \\
        \midrule
        \panet          && R-50-FPN  &&    4.7    &     212.8 &&      36.6 &      58.0 &      39.3 &&      16.3 &      38.1 &      53.1 \\
        \retinamask     && R-101-FPN &&    6.0    &     166.7 &&      34.7 &      55.4 &      36.9 &&      14.3 &      36.7 &      50.5 \\
        \fcis           && R-101-C5  &&    6.6    &     151.5 &&      29.5 &      51.5 &      30.2 &&      8.0  &      31.0 &      49.7 \\
        \mrcnn          && R-101-FPN &&    8.6    &     116.3 &&      35.7 &      58.0 &      37.8 &&      15.5 &      38.1 &      52.4 \\
        \msrcnn         && R-101-FPN &&    8.6    &     116.3 && {\bf 38.3}&      58.8 &      41.5 &&      17.8 &      40.4 &      54.4\\
        \modelname{550} && R-101-FPN && {\bf 33.5}& {\bf 29.8}&&      29.8 &      48.5 &      31.2 &&       9.9 &      31.3 &      47.7  \\
        \midrule
        \modelname{400} && R-101-FPN &&   45.3    &   22.1 &&      24.9 &     42.0 &      25.4 &&       5.0 &      25.3 &      45.0  \\
        \modelname{550} &&  R-50-FPN &&   45.0    &   22.2 &&      28.2 &     46.6 &      29.2 &&       9.2 &      29.3 &      44.8  \\
        \modelname{550} &&  D-53-FPN &&   40.7    &   24.6 &&      28.7 &     46.8 &      30.0 &&       9.5 &      29.6 &      45.5  \\
        \modelname{700} && R-101-FPN &&   23.4    &   42.7 &&      31.2 &     50.6 &      32.8 &&      12.1 &      33.3 &      47.1  \\
        \bottomrule
    \end{smalltable}
    
    \vspace{-0.05in}
    \caption{\inlinesection{MS COCO \cite{coco} Results} We compare to state-of-the-art methods for mask mAP and speed on COCO {\tt test-dev} and include several ablations of our base model, varying backbone network and image size. We denote the backbone architecture with {\tt network-depth-features}, where {\tt R} and {\tt D} refer to ResNet~\cite{resnet} and DarkNet~\cite{yolov3}, respectively. Our base model, \methodname{}-550 with ResNet-101, is 3.9x faster than the previous fastest approach with competitive mask mAP.}
    
    \label{tab:performance}
\end{table*}

%% file: figures/ablations.tex
\begin{table*}
    \centering

    \begin{subfigure}[t]{.36\textwidth}
        \vskip 0pt
        \centering
        \def\mrcnn{Mask R-CNN \cite{maskrcnn}}
        \begin{smalltable}{c r c c c}\toprule
        Method                          &   NMS         &   AP      &  FPS     & Time         \\
        \midrule
        \multirow{2}{*}{\methodname{}}  &   Standard    &{\bf 30.0} & 24.0     & 41.6         \\
                                        &   Fast        & 29.9      &{\bf 33.5}&{\bf 29.8}    \\
        \midrule
        \multirow{2}{*}{Mask R-CNN}     &   Standard    &{\bf 36.1} &  8.6      & 116.0        \\
                                        &   Fast        & 35.8      & {\bf 9.9} &{\bf 101.0}   \\
        \bottomrule
        \end{smalltable}
        \caption{\inlinesection{Fast NMS} Fast NMS performs only slightly worse than standard NMS, while being around 12 ms faster. We also observe a similar trade-off implementing Fast NMS in Mask R-CNN.}
        \label{tab:nms}
    \end{subfigure}
    \,\,\,
    \begin{subfigure}[t]{.21\textwidth}
        \vskip 0pt
        \centering
        \begin{smalltable}{r c c c}\toprule
            $k$ &   AP &  FPS & Time \\
            \midrule
            8 & 26.8 & {\bf 33.0} & {\bf 30.4} \\
            16 & 27.1 & 32.8 & 30.5 \\
            $\vspace{0pt}^*$32  & 27.7 & 32.4 & 30.9 \\
            64  &{\bf 27.8}& 31.7 & 31.5 \\
            128 & 27.6 & 31.5 & 31.8 \\
            256 & 27.7 & 29.8 & 33.6 \\
            \bottomrule
        \end{smalltable}
        \caption{\inlinesection{Prototypes} Choices for $k$. We choose 32 for its mix of performance and speed.}
        \label{tab:num_proto}
    \end{subfigure}
    \,\,\, %
    \begin{subfigure}[t]{.36\textwidth}
        \vskip 0pt
        \centering
        \begin{smalltable}{l c c c}\toprule
            Method                      &   AP      &   FPS     &   Time    \\
            \midrule
            FCIS w/o Mask Voting        &   27.8    &   9.5     &   105.3   \\
            Mask R-CNN (550 $\times$ 550)              &   {\bf 32.2}&   13.5    &   73.9    \\
            \textit{fc}-mask            &   20.7    &   25.7    &   38.9    \\
            \midrule
            \methodname{}-550 (Ours)    &   29.9    & {\bf 33.0}  & {\bf 30.3}   \\
            \bottomrule
        \end{smalltable}
        \caption{\inlinesection{Accelerated Baselines} We compare to other baseline methods by tuning their speed-accuracy trade-offs. \textit{fc}-mask is our model but with $16 \times 16$ masks produced from an \textit{fc} layer.}
        \label{tab:accelerated_baselines}
    \end{subfigure}

    \vspace{-0.1in}   
    \caption{\inlinesection{Ablations} All models evaluated on COCO \texttt{val2017} using our servers. Models in Table \ref{tab:num_proto} were trained for 400k iterations instead of 800k. Time in milliseconds reported for convenience.}
    \vspace{-0.12in}
    \label{tab:ablations}
\end{table*}

%% file: figures/pascal.tex
\begin{table}
    \centering
    
    \def\mnc{MNC \cite{mnc}}
    \def\fcis{FCIS \cite{fcis}}

    \newcommand{\modelname}[1]{\methodname{}-#1}
    
    \begin{smalltable}{l c l cc ccc c ccc cc} \toprule
        Method          & Backbone  &    FPS    &    Time   & $\text{mAP}^r_{50}$ & $\text{mAP}^r_{70}$ \\
        \midrule
        \mnc            &   VGG-16  &    2.8    &      360 &        63.5         &         41.5        \\
        \fcis           & R-101-C5  &    9.6    &      104 &        65.7         &         52.1        \\
        \modelname{550} & R-50-FPN  &{\bf  47.6}&{\bf  21.0}&      {\bf 72.3}     &      {\bf 56.2}     \\
        \bottomrule
    \end{smalltable}
    
    \vspace{-0.05in}
    \caption{\inlinesection{Pascal 2012 SBD \cite{sbd} Results} Timing for FCIS redone on a Titan Xp for fairness. Since Pascal has fewer and easier detections than COCO, YOLACT does much better than previous methods. Note that COCO and Pascal FPS are not comparable because Pascal has fewer classes.}
    
    \label{tab:pascal}
    \vspace{-0.1in}
\end{table}


%% file: figures/more_qualitative.tex
\begin{figure*}
    \centering
    \includegraphics[width=0.9\textwidth]{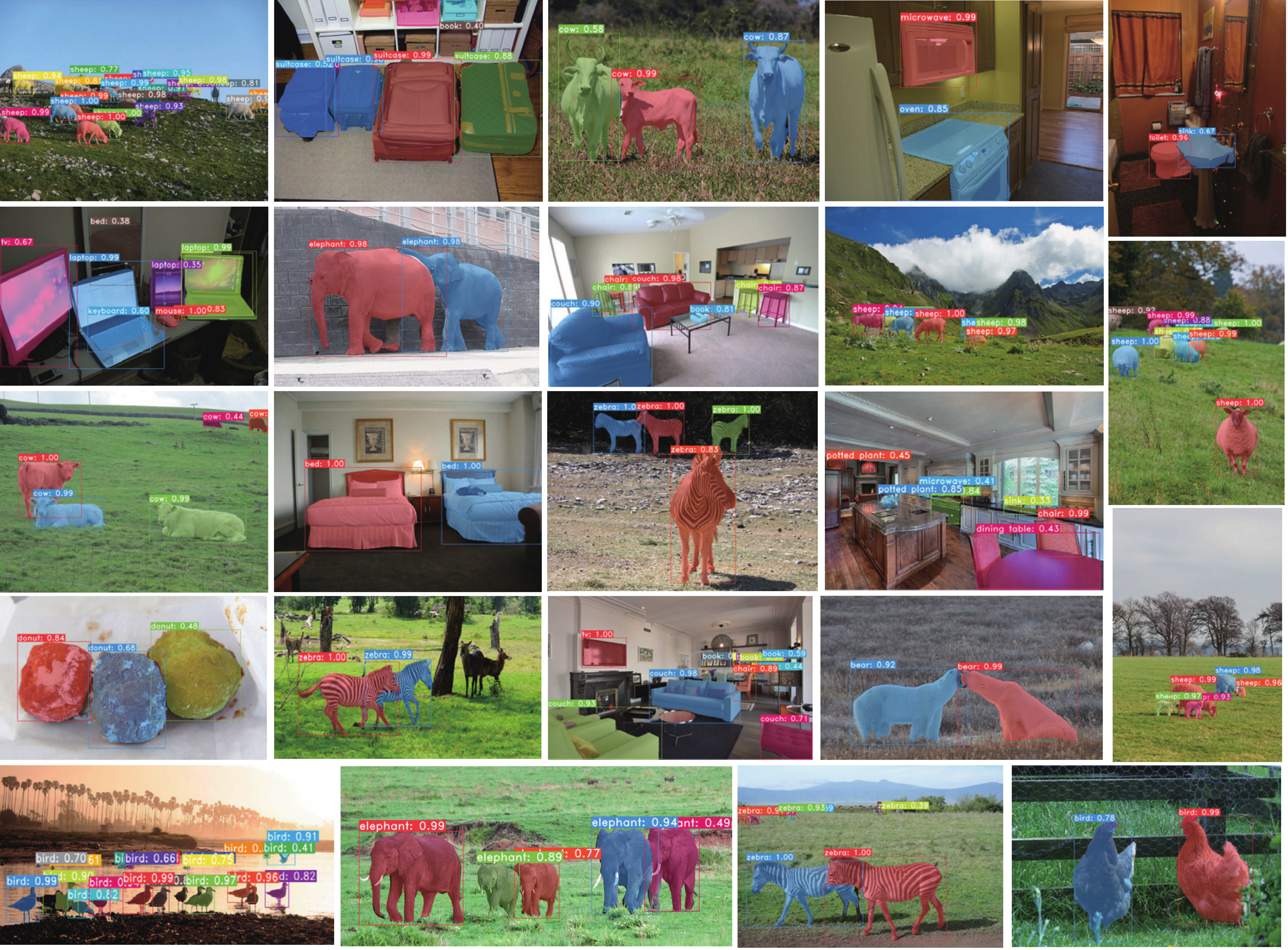}
    \vspace{-0.05in}
    \caption{\textbf{More \methodname{}} evaluation results on COCO's \texttt{test-dev} set with the same parameters as before. To further support that \methodname{} implicitly localizes instances, we select examples with adjacent instances of the same class.}
    \vspace{-0.1in}
    \label{fig:more_qualitative}
\end{figure*}

%% file: figures/detection.tex
\begin{table*}
    \centering
    \newcommand{\yolo}[1]{YOLOv3-#1 \cite{yolov3}}
    \newcommand{\modelname}[1]{\methodname{}-#1}
    
    \begin{smalltable}{l c l cc ccc c ccc cc} \toprule
        Method          && Backbone  &&    FPS                  & Time              &&  AP              & AP$_{50}$ & AP$_{75}$ &&  AP$_{S}$ &  AP$_{M}$ &  AP$_{L}$    \\
        \midrule
        \yolo{320}      && D-53      &&    \textbf{71.17}       & \textbf{14.05}    &&  28.2            & 51.5       & \ph       &&  \ph      &  \ph      &  \ph         \\
        \modelname{400} && R-101-FPN &&    55.43                & 18.04             &&  \textbf{28.4}   & 48.6      & 29.5      &&  10.7     &  28.9     &  43.1        \\
    	\midrule
    	\modelname{550} && R-50-FPN  &&    \textbf{59.30}       & \textbf{16.86}    &&  30.3            & 50.8      & 31.9      &&  14.0     &  31.2     &  43.0        \\
    	\yolo{416}      && D-53      &&    54.47                & 18.36             &&  \textbf{31.0}   & 55.3       & \ph       &&  \ph      &  \ph      &  \ph         \\
        \modelname{550} && D-53-FPN  &&    52.99                & 18.87             &&  \textbf{31.0}   & 51.1      & 32.9      &&  14.4     &  31.8     &  43.7        \\
    	\midrule
        \modelname{550} && R-101-FPN &&    \textbf{41.14}       & \textbf{24.31}    &&  32.3            & 53.0      & 34.3      &&  14.9     &  33.8     &  45.6        \\
    	\yolo{608}      && D-53      &&    30.54                & 32.74             &&  33.0            & 57.9      & 34.4      &&  18.3     &  35.4     &  41.9        \\
    	\modelname{700} && R-101-FPN &&    29.61                & 33.77             &&  \textbf{33.7}   & 54.3      & 35.9      &&  16.8     &  35.6     &  45.7        \\
        \bottomrule
    \end{smalltable}
    \caption{\textbf{\textit{Box} Performance} on COCO's \texttt{test-dev} set. For our method, timing is done without evaluating the mask branch. Both methods were timed on the same machine (using one Titan Xp). In each subgroup, we compare similar performing versions of our model to a corresponding YOLOv3 model. YOLOv3 doesn't report all metrics for the 320 and 416 versions. }
    \label{tab:detection}
\end{table*}